%% file: main.tex
\documentclass{article}

\usepackage{iclr2026_conference,times}
\iclrfinalcopy

\usepackage[utf8]{inputenc}
\usepackage[T1]{fontenc}
\usepackage{amsmath,amssymb,amsfonts}
\usepackage{booktabs}
\usepackage{multirow}
\usepackage{graphicx}
\usepackage{float}
\usepackage{xcolor}
\usepackage{enumitem}
\usepackage{tikz}
\usetikzlibrary{positioning,arrows.meta,calc,fit,backgrounds}
\usepackage{pgfplots}
\pgfplotsset{compat=1.18}
\usepgfplotslibrary{groupplots}
\usepackage{hyperref}
\hypersetup{colorlinks=true, linkcolor=blue!45!black, citecolor=blue!45!black, urlcolor=blue!45!black}

\definecolor{sdcteal}{HTML}{1D9E75}
\definecolor{sdctealdark}{HTML}{0F6E56}
\definecolor{gdngray}{HTML}{888780}
\definecolor{barneutral}{HTML}{B4B2A9}
\definecolor{fullblue}{HTML}{378ADD}
\definecolor{coral}{HTML}{D85A30}
\definecolor{gridgray}{HTML}{D3D1C7}

\newcommand{\method}{HOLA}
\newcommand{\be}{\beta\!\cdot\!\lVert e\rVert}     
\newcommand{\norm}[1]{\lVert #1 \rVert}
\newcommand{\T}{^{\!\top}}
\newcommand{\ostate}{o^{\mathrm{state}}}
\newcommand{\ocache}{o^{\mathrm{cache}}}

\title{\bfseries A Hippocampus for Linear Attention\\[3pt]
  {\mdseries\large An Exact Memory for What the Recurrent State Forgets}}
\author{Wanyun Cui \\ Shanghai University of Finance and Economics}
\date{}

\begin{document}
\maketitle
\lhead{} 

\begin{abstract}
Linear-attention and state-space language models compress the prefix into a fixed-size recurrent state,
yielding $O(1)$ memory at the cost of a lossy exact memory: when many key--value associations compete, earlier facts are
overwritten and needle recall degrades. Inspired by Complementary Learning Systems, we give linear attention a
hippocampal complement. \method{} (Hippocampal Linear Attention) keeps the usual delta-rule state as a
compressive memory and adds a bounded exact KV cache, forming a semiparametric test-time memory: the state
models linearly compressible structure, while the cache stores associations that should not be forced through
that state. The cache writes without a learned eviction module, keeping tokens with large $\be$, the
prediction residual actually committed to the state; a decoupled RMSNorm-$\gamma$ cache read then turns these
exact KV pairs into sharp retrieval rather than soft averaging. At 340M parameters trained on 15B SlimPajama
tokens, \method{} lowers Wikitext perplexity from $27.32$ to $22.92$ ($-16.1\%$), below a full-attention
Transformer++ ($26.88$), and improves LAMBADA perplexity from $30.95$ to $30.26$. It also achieves the best
linear in-context retrieval and remains much more robust than GDN or a matched \method{}+recency cache on RULER
needle-in-a-haystack recall out to 32k tokens ($16\times$ its training length).
\end{abstract}

\begin{figure}[H]
\vspace{-0.5em}
\centering
\resizebox{\textwidth}{!}{\input{fig_teaser}}
\vspace{-0.7em}
\caption{\textbf{\method{} lowers perplexity and improves length-robust needle recall.} \textbf{(a)} At 340M,
\method{} reduces Wikitext perplexity from $27.3$ to $22.9$ ($-16.1\%$), below full-attention Transformer++
($26.9$). \textbf{(b)} On RULER S-NIAH-1, \method{} remains stronger than GDN and \method{}+recency as
context grows to 32k tokens.}
\label{fig:teaser}
\end{figure}
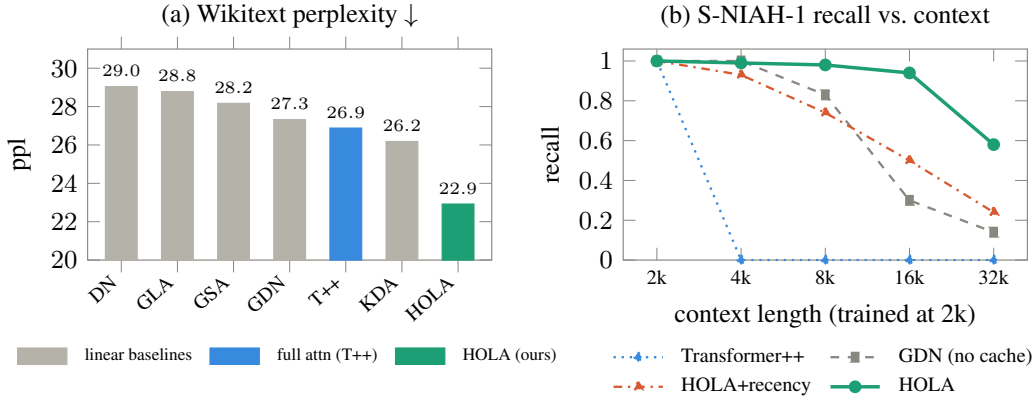

\input{sec_intro}
\input{sec_related}
\input{sec_method}
\input{sec_experiments}
\input{sec_conclusion}

\bibliographystyle{iclr2026_conference}
\bibliography{references}

\appendix
\input{sec_appendix}

\end{document}

%% file: fig_teaser.tex
\begin{tikzpicture}
\begin{groupplot}[
  group style={group size=2 by 1, horizontal sep=1.6cm},
  width=6.5cm, height=4.15cm,
  tick label style={font=\footnotesize}, label style={font=\footnotesize},
  title style={font=\small, yshift=-2pt},
  every node near coord/.append style={font=\tiny, /pgf/number format/fixed,
       /pgf/number format/precision=1, /pgf/number format/fixed zerofill},
  axis line style={gdngray}, tick style={gdngray},
]
\nextgroupplot[
  title={(a) Wikitext perplexity $\downarrow$},
  ybar, bar width=11pt,
  xtick={1,2,3,4,5,6,7}, xticklabels={DN,GLA,GSA,GDN,T++,KDA,\method},
  x tick label style={rotate=42,anchor=east,font=\scriptsize},
  xmin=0.4, xmax=7.6,
  ymin=20, ymax=31, ytick={20,22,24,26,28,30},
  ylabel={ppl},
  nodes near coords,
  area legend,
  legend style={font=\tiny, draw=none, at={(0.5,-0.36)}, anchor=north, legend columns=3,
                column sep=4pt},
  legend cell align=left,
]
\addplot[ybar, bar shift=0pt, fill=barneutral, draw=barneutral]
  coordinates {(1,29.04) (2,28.78) (3,28.17) (4,27.32) (6,26.18)};
\addlegendentry{linear baselines}
\addplot[ybar, bar shift=0pt, fill=fullblue, draw=fullblue]
  coordinates {(5,26.88)};
\addlegendentry{full attn (T++)}
\addplot[ybar, bar shift=0pt, fill=sdcteal, draw=sdcteal]
  coordinates {(7,22.92)};
\addlegendentry{\method{} (ours)}
\nextgroupplot[
  title={(b) S-NIAH-1 recall vs.\ context},
  symbolic x coords={2k,4k,8k,16k,32k}, xtick=data,
  x tick label style={font=\scriptsize},
  xlabel={context length (trained at 2k)},
  ymin=0, ymax=1.06, ytick={0,0.2,0.4,0.6,0.8,1.0},
  ylabel={recall}, enlarge x limits=0.10,
  legend style={font=\scriptsize, at={(0.5,-0.36)}, anchor=north, legend columns=2, draw=none,
                column sep=4pt},
  legend cell align=left,
]
\addplot[fullblue, thick, dotted, mark=diamond*, mark options={fill=fullblue,scale=0.85}]
  coordinates {(2k,1.00)(4k,0.00)(8k,0.00)(16k,0.00)(32k,0.00)};
\addplot[gdngray, thick, dashed, mark=square*, mark options={fill=gdngray,scale=0.75}]
  coordinates {(2k,1.00)(4k,1.00)(8k,0.83)(16k,0.30)(32k,0.14)};
\addplot[coral, thick, dash dot, mark=triangle*, mark options={fill=coral,scale=0.85}]
  coordinates {(2k,1.00)(4k,0.93)(8k,0.74)(16k,0.50)(32k,0.24)};
\addplot[sdcteal, very thick, mark=*, mark options={fill=sdcteal,scale=0.8}]
  coordinates {(2k,1.00)(4k,0.99)(8k,0.98)(16k,0.94)(32k,0.58)};
\legend{Transformer++, GDN (no cache), \method{}+recency, \method{}}
\end{groupplot}
\end{tikzpicture}

%% file: sec_intro.tex
\section{Introduction}
\label{sec:intro}

The human brain does not rely on a single memory. A hippocampus-centered system can record a specific, novel
event in one shot and later recall it precisely~\citep{tse2007science}, while a neocortex-centered system
slowly distills generalizable structure across many experiences~\citep{frankland2005nrn}; Complementary
Learning Systems (CLS) theory~\citep{mcclelland1995cls,kumaran2016cls} argues that the two \emph{must} be
separate --- a system built for slow, compressive generalization would suffer \emph{catastrophic
interference} if forced to write single facts quickly, and so cannot also serve fast, exact, one-shot memory.
Efficient language models are missing exactly this complementarity.

\textbf{Linear attention is precisely such a ``neocortex'': a superb compressor, but a leaky exact memory.}
DeltaNet, Gated DeltaNet (GDN), and related linear-attention / state-space models
\citep{katharopoulos2020linear,gu2023mamba,yang2024gdn,yang2024deltanet} replace softmax attention's growing
key--value cache with a fixed-size recurrent state $S$, summarizing all history in $O(1)$ memory and
processing a sequence in sub-quadratic time. The state is updated online --- the \emph{delta rule} writes the
residual between each new value and the state's current prediction --- making it an efficient running summary.
But a fixed-size state is a \emph{lossy} memory: it holds only a bounded number of distinct key$\to$value
associations before new writes overwrite old ones along shared key directions. This shows up exactly where it
hurts: multi-item associative recall~\citep{arora2024zoology}, passkey / needle-in-a-haystack
retrieval~\citep{mohtashami2023landmark}, and copying a specific distant token
verbatim~\citep{jelassi2024copying}. There, full softmax attention --- which
keeps every token exactly, at $O(T)$ memory and $O(T^2)$ compute --- remains the gold standard. The question
becomes: can we keep linear attention's cost while recovering its lost exact recall? CLS points to the answer
 --- give this neocortex a hippocampus: a separate, exact, budget-limited episodic memory that preferentially
stores \emph{surprise}. We formalize this complement as \emph{semiparametric test-time memory regression}:
the recurrent state is the parametric estimator for linearly compressible structure, and a bounded exact cache
is the non-parametric correction over KV pairs the state should not be forced to absorb. This framework
dictates two design choices: \emph{what to store} (the surprising items) and \emph{how to read it} (exact
retrieval, not soft averaging).

\textbf{What to store: a surprise signal the model already computes.} A budget-limited episodic memory is only
useful if it keeps the \emph{right} tokens --- the surprising ones, which the compressed state could not
absorb. The obvious choice is instead to keep the most \emph{recent} tokens, a sliding window, as most
linear-attention hybrids do \citep{xiao2023streamingllm,zhang2023h2o,nha2025,rattention2025,ahn2025}; but recency
cannot represent information that is far away yet still must be recalled exactly --- once it slides out of the
window it is as lost as if the state had overwritten it. Our key observation is that a delta-rule model already
measures how surprising each token is to its state: if the state predicts the token's value poorly and the
model strongly commits the correction, that token is exactly the kind of support point the non-parametric cache
should keep. We use this intrinsic write magnitude as the cache's eviction score, retaining tokens that changed
the state most rather than tokens that are merely recent. A matched control confirms the intended separation:
at 340M, importance eviction beats a recency cache on perplexity and long-context retrieval, while commonsense
remains within single-seed noise (Sec.~\ref{sec:exp}).

\textbf{How to read: retrieval, not averaging.} An exact copy buys nothing if it is read like another linear
summary. Directly reusing the L2-normalized queries and keys used by DeltaNet/GDN makes cache logits too small,
so the cache softmax is nearly uniform and the exact memory degenerates into a lossy average. A simple fix is
to use Qwen3-style RMSNorm-$\gamma$ \citep{yang2025qwen3} on the cache path only, decoupled from the state
update. This restores sharp, near-argmax retrieval while preserving the unit-norm keys required for the delta
rule's stable state update.

Together, these ideas give \textbf{Hippocampal Linear Attention (HOLA)}, which recovers the exact recall the
state loses (Fig.~\ref{fig:teaser}). At 340M parameters trained on 15B SlimPajama tokens, the model reaches the
lowest perplexity of every model we compare --- \emph{below} a full-attention Transformer++ (Wikitext $22.9$
vs.\ $26.9$) --- and, unlike GDN, stays length-robust on hard RULER recall out to 32k tokens ($16\times$ its
2k training length), where the recurrent state saturates and GDN decays.

\paragraph{Contributions.}
\begin{enumerate}[leftmargin=1.4em,itemsep=2pt,topsep=2pt]
\item \textbf{A hippocampal exact memory for linear attention, motivated by CLS.} We argue that a fixed
recurrent state is a neocortex-like compressor but a leaky exact memory, and add a bounded, hippocampus-like KV
store for precise one-shot associations.
\item \textbf{A semiparametric test-time memory framework.} We formalize memory readout as test-time regression
over prefix key--value observations: pure GDN is the parametric estimator $q\T S_t$, full attention is an
unbounded non-parametric kernel estimator, and \method{} is the bounded semiparametric case.
\item \textbf{A concrete write/read algorithm.} The cache writes by an intrinsic surprise signal already
computed by the delta rule --- the committed residual update to the state --- and reads by a decoupled
RMSNorm-$\gamma$ cache path that makes exact KV pairs retrievable rather than softly averaged.
\item \textbf{Large empirical gains in perplexity and retrieval.} At 340M, \method{} sharply lowers Wikitext
perplexity relative to the same-backbone GDN anchor ($27.32\!\to\!22.92$), falls below a full-attention
Transformer++ ($26.88$), achieves the best linear in-context retrieval, and remains much more robust than GDN
or a recency cache on long-context needle recall.
\end{enumerate}

%% file: sec_related.tex
\section{Related Work}
\label{sec:related}

Our work connects three threads: linear-time recurrent attention, hybrid exact-memory mechanisms for efficient
language models, and semiparametric memory.

\paragraph{Linear-time recurrent attention.}
Linear attention \citep{katharopoulos2020linear}, state-space models
\citep{gu2023mamba,dao2024mamba2}, and the DeltaNet family
\citep{schlag2021fwp,yang2024deltanet} replace softmax attention's growing KV cache with a fixed-size
recurrent state. Gated DeltaNet (GDN) \citep{yang2024gdn} strengthens DeltaNet with data-dependent decay, and
GLA \citep{yang2024gla}, GSA \citep{zhang2024gsa}, and Kimi Delta Attention (KDA) \citep{kimi2025linear}
are representative modern linear-attention baselines. This fixed state is efficient but lossy: as an associative memory, it has bounded
capacity before new writes interfere with old ones \citep{hopfield1982,ramsauer2021hopfield}. Based and
Zoology \citep{arora2024based,arora2024zoology} make this recall--throughput trade-off explicit, showing that
efficient recurrent models lag softmax attention on multi-item associative recall. \method{} keeps the
recurrent backbone, but adds a bounded exact memory for the KV pairs the state should not be forced to
compress.

\paragraph{Hybrid exact memory for efficient LMs.}
Many efficient LMs recover some exact recall by pairing a recurrent/SSM backbone with softmax attention.
Inter-layer hybrids such as Jamba \citep{lieber2024jamba}, Samba \citep{ren2024samba}, and Griffin
\citep{de2024griffin} mix recurrent layers with local attention layers; layer-internal hybrids such as NHA
\citep{nha2025}, RAttention \citep{rattention2025}, and AHN \citep{ahn2025} combine recurrent memory with a
recent-token window or learned compressed store. These designs differ architecturally, but their lossless
component is primarily recency-based: once an old token leaves the window, it must be represented by the
compressed state. \method{} instead makes the exact memory selective over the prefix, retaining surprising
KV pairs even when they are far from the current query.

The closest work is LTE \citep{lte2025}, which also augments GDN with a bounded evictable KV cache. The main
difference is how the cache is selected and integrated: LTE learns eviction scores with an additional CNN
module and alternates GDN with sparse-attention layers, whereas \method{} uses the delta rule's own write
magnitude as a parameter-free surprise score and attaches the cache inside every recurrent layer. Thus LTE
shows that evictable caches can help delta-rule models, while \method{} argues that the recurrent update itself
already exposes the right signal for what an exact memory should keep.

\paragraph{Semiparametric and retrieval-augmented memory.}
Semiparametric language models pair parametric sequence modeling with explicit non-parametric memory, as in
kNN-LM \citep{khandelwal2020knnlm}, RETRO \citep{borgeaud2022retro}, Memorizing Transformers
\citep{wu2022memorizing}, and SPALM \citep{yogatama2021spalm}. Those systems retrieve from an external or
long-lived datastore. Our setting is in-sequence and test-time: the parametric part is the recurrent state
$S_t$, the non-parametric part is a bounded set of exact in-context KV pairs, and the readout interpolates
between the two. This semiparametric view also places softmax attention and GDN on the same spectrum: full
attention is an unbounded non-parametric estimator over all prefix tokens, while GDN is the purely parametric
state estimator.

%% file: sec_method.tex
\section{Method}
\label{sec:method}

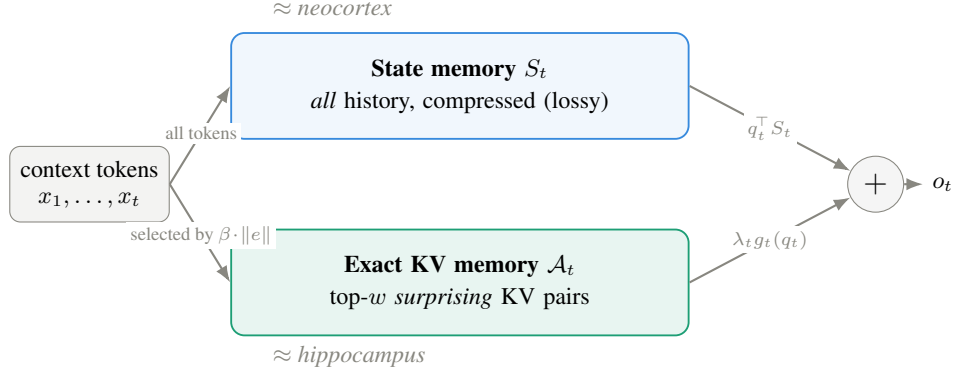
\begin{figure}[H]
\centering
\input{fig_method}
\caption{\textbf{\method{}: semiparametric test-time memory.} Every token updates the recurrent
\emph{state memory} ($\approx$ neocortex; lossy, $O(1)$), while the tokens with large delta-rule write magnitude
$\be$ are additionally kept as exact KV pairs in a bounded exact-KV memory ($\approx$ hippocampus). The
read-out follows the semiparametric form $o_t=q_t\T S_t+\lambda_t g_t(q_t)$: a compressive state estimate plus
a non-parametric exact-KV read, instantiated with a sharpened softmax in Sec.~\ref{sec:method:read}.}
\label{fig:method}
\end{figure}

We derive \method{} from the delta-rule update equation: a single equation tells us both why a cache is
necessary (Sec.~\ref{sec:method:framework}) and \emph{what to store in it} (Sec.~\ref{sec:method:store}); we
then address \emph{how to read it} (Sec.~\ref{sec:method:read}) and the full layer
(Sec.~\ref{sec:method:impl}). Figure~\ref{fig:method} gives an overview.

\subsection{Background: the delta rule and its lossiness}
\label{sec:method:bg}
DeltaNet compresses history into a $(d_k,d_v)$ matrix $S$, an \emph{online associative memory}: the read-out
for key $k$, $k\T S$, gives its currently associated value. Let token $t$ have query/key/value $q_t,k_t,v_t$
(with $q_t,k_t$ L2-normalized to unit norm) and write strength $\beta_t\in[0,1]$. The delta rule updates by
``predict, then write the residual'':
\begin{equation}
S_t \;=\; S_{t-1} + \beta_t\, k_t\, e_t\T,
\qquad e_t \;=\; v_t - k_t\T S_{t-1},
\qquad \ostate_t \;=\; q_t\T S_t .
\label{eq:deltarule}
\end{equation}
Here $e_t$ is the \emph{residual} (innovation): the part of $v_t$ that $S$ cannot already predict along
$k_t$. Writing the residual (rather than $v_t$) suppresses repeated same-key interference.

\textbf{It is lossy.} $S$ is fixed-size, of rank at most $d_k$: once the number of distinct associations
exceeds capacity, new writes \emph{overwrite} old ones along shared key directions, and a distant $(k,v)$ can
no longer be recovered exactly from $S$. This is the ``neocortex-like'' deficiency of
Sec.~\ref{sec:intro} --- good at generalizing, unable to keep distant exact detail. The next three subsections
build \method{} on top; the specific backbone we use (GDN) is deferred to Sec.~\ref{sec:method:impl}.

\subsection{Semiparametric memory as test-time regression}
\label{sec:method:framework}
The previous subsection shows why a fixed recurrent state is lossy. We now give the framework that turns this
observation into a design principle: memory readout is a test-time regression problem, and \method{} is a
bounded semiparametric estimator for it.

\begin{quote}
\noindent\textbf{Definition 1 (test-time memory regression).}
\itshape
For a layer at position $t$, let the causally available key--value observations be
$\mathcal{D}_t=\{(k_i,v_i)\}_{i\le t}$. A test-time memory regression (TMR) mechanism is a memory state
$\mathcal{M}_t$ together with two operations
\[
\mathcal{M}_t=\mathrm{Write}(\mathcal{M}_{t-1},k_t,v_t),
\qquad
 o_t=\mathrm{Read}(q_t,\mathcal{M}_t)=\hat f_t(q_t),
\]
where $\hat f_t$ is an estimator, built from $\mathcal{D}_t$, of the context-specific map $f_t:q\mapsto v$.
The choice of memory state $\mathcal{M}_t$ and estimator class $\hat f_t$ specifies the memory mechanism.
\end{quote}

Definition~1 makes the comparison clean. GDN is a TMR mechanism whose memory state is only a fixed-size matrix
$\mathcal{M}_t=S_t$ and whose read operation is linear,
\begin{equation}
o_t=\mathrm{Read}(q_t,S_t)=\hat f_{\mathrm{state},t}(q_t)=q_t\T S_t .
\end{equation}
Here $S_t$ is a \emph{test-time} parameter: it is updated online from the context by the delta rule, not learned
as an additional model weight. It is cheap and captures the linearly compressible part of the key--value map,
but it cannot interpolate all observations in $\mathcal{D}_t$ once the context exceeds its capacity.

\begin{quote}
\noindent\textbf{Definition 2 (semiparametric TMR).}
\itshape
A TMR mechanism from Definition~1 is semiparametric if its memory decomposes as
$\mathcal{M}_t=(S_t,\mathcal{A}_t)$, where $S_t$ is a fixed-size parametric state and
$\mathcal{A}_t\subseteq\mathcal{D}_t$ is a set of exact KV pairs kept non-parametrically. Its Write operation
updates both components,
\[
(S_t,\mathcal{A}_t)=\mathrm{Write}((S_{t-1},\mathcal{A}_{t-1}),k_t,v_t),
\]
where the state update changes the parametric estimator and the KV-set update may admit or evict exact KV
pairs. Its Read operation returns
\begin{equation}
o_t = \mathrm{Read}(q_t,\mathcal{M}_t)=\hat f_t(q_t)
     = q_t\T S_t + \lambda_t\,g_t(q_t).
\label{eq:combine}
\end{equation}
Here $\lambda_t$ is a read-side mixing coefficient, and $g_t$ is a non-parametric model over the exact KV pairs
in $\mathcal{A}_t$.
\end{quote}

This form also locates full softmax attention. It is the non-parametric TMR instance obtained by disabling the
state term and using all causally available observations as exact KV pairs, i.e., set $\mathcal{A}_t=\mathcal{D}_t$. In
that case the non-parametric model $g_t$ is exactly the softmax kernel estimator,
\begin{equation}
g_{\mathrm{attn},t}(q)=
\sum_{(k_i,v_i)\in\mathcal{D}_t}
\frac{\exp(q\T k_i/\sqrt d)}{\sum_{(k_j,v_j)\in\mathcal{D}_t}\exp(q\T k_j/\sqrt d)}v_i,
\end{equation}
i.e.\ a Nadaraya--Watson kernel estimator \citep{nadaraya1964estimating,watson1964smooth}. It can recover
exact items because every token remains an exact KV pair, but the KV set grows with context length.

\method{} is the bounded semiparametric TMR instance. We take $S_t$ to be the full GDN recurrent state and keep
a bounded set of exact KV pairs selected from $\mathcal{D}_t$ by the score in Sec.~\ref{sec:method:store}. Under
this framework, \method{} has two design questions: which exact KV pairs to keep
(Sec.~\ref{sec:method:store}), and how sharp the kernel read should be (Sec.~\ref{sec:method:read}).

\subsection{What to store: the write magnitude $\be$ is ``surprise''}
\label{sec:method:store}
We should store the tokens that are least well represented by the state memory. The \emph{same update
equation already says which}. Write the update as
\begin{equation}
S_t = S_{t-1} + \Delta_t,\qquad \Delta_t = \beta_t\, k_t e_t\T \quad(\text{rank-1}).
\end{equation}
The token's entire effect on $S$ is this one rank-1 matrix $\Delta_t$. How much a token ``writes'' is
naturally the size of $\Delta_t$; for a rank-1 matrix the Frobenius norm factorizes, and with
$\norm{k_t}=1$ collapses to a scalar:
\begin{equation}
m_t \;=\; \norm{\Delta_t}_F \;=\; \beta_t\, \norm{k_t}\, \norm{e_t} \;=\; \beta_t\, \norm{e_t}.
\label{eq:score}
\end{equation}
We write this as $\be$ and use it directly as the eviction score. Its meaning is immediate: \textbf{$m_t$ is
how much the token changed $S$.} A large change means the token brought information $S$ could \emph{not}
predict (large residual $e_t$, i.e.\ the innovation in the Kalman~\citep{kalman1960} sense; the delta rule is
the Widrow--Hoff/LMS rule~\citep{widrow1960}, where $e$ is its error term) and that the model actually wrote it in
(large $\beta_t$). A token with small residual is already well predicted by the state, so storing a verbatim
copy is wasteful; a token with large committed residual is exactly where the compressed state needed the most
help. \emph{In short, we spend the limited exact memory where the state-memory representation is weakest.}

\textbf{Cache $=$ the top-$w$ tokens seen so far by $\be$.} Each layer keeps a persistent exact cache of
capacity $w$ (default $w{=}64$). Its members are the exact $(k,v)$ copies of the $w$ highest-$m$ tokens in the
causal history observed so far -- keep what wrote most to $S$, \emph{regardless of distance}. Because $m_t$ is
fixed when the token is written, the same top-$w$ set can be maintained online or selected blockwise without
order dependence; training and inference therefore use the same cache semantics.

\textbf{Versus recency.}\quad Sliding-window memories, as used in StreamingLLM~\citep{xiao2023streamingllm},
NHA~\citep{nha2025}, and RAttention~\citep{rattention2025}, choose exact KV pairs by position. This is
useful for local context, but it cannot keep an old item solely because it remains important. Selection by
$\be$ instead keeps surprising KV pairs across distance, compensating the state-memory failure described
in Sec.~\ref{sec:method:bg}. An ablation (Sec.~\ref{sec:exp}) shows why the product matters: the residual
alone ($\norm{e}$) or the write strength alone ($\beta\norm{v}$) each underperform; their product is best.

For the read in Sec.~\ref{sec:method:read}, we denote the bounded visible KV set by $\mathcal{V}_t$. Its main
persistent component is the top-$w$ exact cache above; in implementation we also include the causal tokens in
the current processing block and one null sink. These additions are bounded bookkeeping for causal block
processing and stability; the persistent exact memory is selected by $\be$.

\subsection{How to read: retrieval, not soft averaging}
\label{sec:method:read}
An exact copy is worthless if \emph{read} like a linear attention. If the cache reuses the backbone's
unit-L2-normalized $q,k$ \citep{yang2024deltanet,yang2024gdn}, the effective logit is
$\tau\cdot(1/\sqrt{d})\cos \approx 0.83\cos$ (a learned
$\tau\!\approx\!6.6$, head dimension $d$), ranging only over $\pm0.83$: the softmax is nearly uniform --- among
$w{=}64$ entries a perfectly matching key receives only ${\sim}3.5\%$ of the mass. The cache degenerates into
yet another soft-averaging lossy summary (explaining why a naive cache barely helps).

\textbf{Sharpening via a decoupled RMSNorm-$\gamma$.} We apply Qwen3-style RMSNorm with a learnable $\gamma$ to
the \emph{cache-path} $q,k$ (keeping the norm at $\sqrt{d}\!\approx\!11$ rather than $1$) and fix $\tau{=}1$. The
cache read is then a sharpened softmax attention over $\mathcal{V}_t$:
\begin{equation}
\ocache_t \;=\; \sum_{j\in\mathcal{V}_t}
\mathrm{softmax}_j\!\Big(\tilde q_t\T \tilde k_j / \sqrt{d}\Big)\, v_j,
\qquad \tilde q = \mathrm{RMSNorm}_\gamma(q),\ \ \tilde k = \mathrm{RMSNorm}_\gamma(k).
\label{eq:read}
\end{equation}
Since $\norm{\tilde q}\!\approx\!\norm{\tilde k}\!\approx\!\sqrt{d}$, the effective logit is
${\approx}\sqrt{d}\cos\approx 11\cos$ (vs.\ $0.83\cos$ for unit-L2), so the cache finally performs
\emph{near-argmax retrieval}. This change \emph{acts only on the cache read and is decoupled from the
state-update path} --- the $q,k$ feeding $S$ remain unit-L2-normalized. Decoupling is necessary: the delta rule
relies on $\norm{k}=1$ to keep the update operator $I-\beta\,kk\T$ within $[0,1]$ eigenvalues (stable); a
$\sqrt{d}$ norm in the state update would give eigenvalues $1-\beta d$ and diverge. The learnable $\gamma$
self-moderates sharpness: versus a fixed high temperature it reaches lower perplexity while avoiding ``the
state grows lazy and far recall collapses'' (Sec.~\ref{sec:exp}). Empirically, \textbf{sharpening is the
single largest lever in the design} (perplexity $70\!\to\!60$, ${\sim}2\times$ multi-key capacity).

\subsection{Instantiation}
\label{sec:method:impl}
\textbf{Instantiation (GDN).} The derivation used only the delta rule. We instantiate on the strongest
linear backbone, Gated DeltaNet, which adds a data-dependent decay gate $\alpha_t\in(0,1]$ (selective
forgetting), making the prediction $\alpha_t k\T S_{t-1}$ and the residual
$e_t=v_t-\alpha_t k\T S_{t-1}$. The gate is \emph{orthogonal} to our method and $\be$ is unchanged; all
experiments use GDN.

\textbf{Overhead over GDN.} \method{} is almost iso-parametric with its GDN backbone. At 340M
($L{=}24$, $H{=}4$, $d{=}256$), the only learned cache-specific parameters are the cache-path Q/K
RMSNorm scales plus a per-head sink and cache gate:
$L(2d+2H)=24(512+8)=12{,}480$ trainable scalars, less than $0.004\%$ of the full model
(the frozen temperature adds only $LH{=}96$ stored scalars). The cache itself is inference state rather than
model weights: in bf16 decoding it stores at most $(w{+}C)$ K/V pairs per layer, about
$24\cdot320\cdot4\cdot256\cdot2\cdot2\approx31$ MB plus negligible scores. Measured peak GPU allocation
(weights included, bs$=1$ decode) is therefore close to GDN and flat with context: $0.75$ GB for \method{}
versus $0.72$ GB for GDN at both 32k and 128k. Thus the gains do not come from a larger parametric model, and
the exact-memory overhead over GDN is small ($\approx5\%$ peak memory).

%% file: fig_method.tex
\begin{tikzpicture}[
  font=\small,
  mem/.style={rectangle, rounded corners=5pt, draw, align=center, text width=5.7cm,
              minimum height=1.4cm, inner sep=5pt, line width=0.7pt},
  state/.style={mem, draw=fullblue, fill=fullblue!7},
  exact/.style={mem, draw=sdcteal, fill=sdcteal!10},
  io/.style={rectangle, rounded corners=4pt, draw=gdngray, fill=gdngray!10, align=center,
             minimum height=1.0cm, inner sep=4pt},
  sumc/.style={circle, draw=gdngray, fill=gdngray!8, minimum size=0.74cm, inner sep=0pt},
  ar/.style={-{Latex[length=2.2mm]}, gdngray, line width=0.8pt},
  tag/.style={font=\footnotesize\itshape, gdngray},
  elab/.style={font=\scriptsize, gdngray, fill=white, inner sep=1.5pt},
]
\node[io] (x) at (0,-0.7) {context tokens\\ $x_1,\dots,x_t$};
\node[state] (S) at (4.9,0.6)  {\textbf{State memory}~$S_t$\\[2pt]{\footnotesize \emph{all} history, compressed (lossy)}};
\node[exact] (E) at (4.9,-2.0) {\textbf{Exact KV memory}~$\mathcal{A}_t$\\[2pt]{\footnotesize top-$w$ \emph{surprising} KV pairs}};
\node[tag, anchor=west] at (2.3,1.62)  {$\approx$ neocortex};
\node[tag, anchor=west] at (2.3,-2.98) {$\approx$ hippocampus};
\node[sumc] (sum) at (10.4,-0.7) {\large$+$};
\node[right=2.5mm of sum] (o) {$o_t$};

\draw[ar] (x.east) -- (S.west) node[elab, pos=0.52] {all tokens};
\draw[ar] (x.east) -- (E.west) node[elab, pos=0.52] {selected by~$\be$};
\draw[ar] (S.east) -- (sum)    node[elab, pos=0.5] {$q_t\T S_t$};
\draw[ar] (E.east) -- (sum)    node[elab, pos=0.5] {$\lambda_t g_t(q_t)$};
\draw[ar] (sum.east) -- (o);
\end{tikzpicture}

%% file: sec_experiments.tex
\section{Experiments}
\label{sec:exp}

\subsection{Setup}
\textbf{Architecture.} We use the GDN architecture at 340M~\citep{yang2024gdn}: $d_{\mathrm{model}}{=}1024$, 24 layers,
$4$ heads $\times$ head-dim $256$, $\mathrm{expand\_v}{=}1$, $\mathrm{hidden\_ratio}{=}4$, conv 4, tied
embeddings, and vocabulary $32000$. \method{} adds a per-layer cache: window $w{=}64$, chunk $C{=}256$,
eviction score $\be$, cache normalization RMSNorm-$\gamma$, $\tau{=}1$ frozen, gate init $-4$. The GDN baseline
and \method{} \emph{share an identical backbone}; the only difference is the cache, so internal comparisons are
strictly controlled.

\textbf{Training recipe.} We follow the Preconditioned-DeltaNet 340M recipe~\citep{precond2026}: SlimPajama
15.0B tokens \citep{soboleva2023slimpajama}, Mistral tokenizer \citep{jiang2023mistral}, context 2048; AdamW
\citep{loshchilov2019adamw} (peak lr $4{\times}10^{-4}$, wd $0.01$, cosine, warmup 1000, grad-clip 1.0), batch
0.5M tokens, 1 epoch. This recipe match enables reuse of the published DeltaNet/KDA/GDN baseline rows. Trained
on $8\times$A800.

\textbf{Baselines.} We train our own GDN anchor and \method{} for the controlled same-backbone comparison, and
borrow recipe-matched architecture rows for Transformer++, GLA, and GSA from \citet{nha2025}, and DeltaNet, KDA,
and GDN from \citet{precond2026}; KDA itself is from Kimi Linear~\citep{kimi2025linear}.

\textbf{Evaluation.} (1) Language modeling: Wikitext-103 perplexity, LAMBADA. (2) Commonsense (zero-shot):
ARC-e/c, PIQA, HellaSwag, WinoGrande, BoolQ, SciQ, OpenBookQA, LAMBADA-acc. (3) In-context retrieval: FDA,
SWDE, SQuAD. (4) Long context: RULER \citep{hsieh2024ruler} (2k$\to$32k; multi-key MK, multi-value MV,
multi-query MQ, variable tracking VT) and passkey/needle.

\subsection{Main results: language modeling, commonsense, retrieval}
Table~\ref{tab:main} reports the main comparison. The effect is large on the two axes the cache is designed to
improve: \textbf{\method{} sharply lowers perplexity and substantially boosts in-context retrieval}. Wikitext
perplexity drops from $27.32\!\to\!22.92$ relative to our same-backbone GDN anchor (${-}16.1\%$), and even falls
below the full-attention Transformer++ ($26.88$). Retrieval improves just as strongly: FDA rises
$11.7\!\to\!20.1$ ($+72\%$ relative) and SWDE $29.0\!\to\!35.9$ ($+24\%$), the best among linear models, while
commonsense remains competitive on the six-task average.

\begin{table}[t]
\centering
\caption{340M / SlimPajama-15B / ctx-2048 main comparison ($\downarrow$/$\uparrow$ lower/higher better).
Commonsense is broken out per task (ARCe, PIQA, Wino, LMBa $=$ accuracy; ARCc, Hella $=$ acc\_norm), and
Avg. is their six-task mean. Borrowed rows are from recipe-matched papers (Transformer++/GLA/GSA \citep{nha2025};
DeltaNet/GDN \citep{precond2026}; KDA is Kimi Delta Attention~\citep{kimi2025linear}, with numbers borrowed from \citealp{precond2026}); their per-task values reproduce each source's reported commonsense average exactly.
Bold $=$ best per column among the sub-quadratic models (Transformer++ excluded, as it is a different,
quadratic-cost class). ``\method{}${+}$recency'' is our own matched position-eviction variant (HOLA's cache but with recency instead of $\be$ eviction), with
retrieval via the corrected full-recompute evaluation path. Commonsense differences among our three models (GDN / \method{}${+}$recency /
\method{}) are within single-seed noise (${<}0.7$ on the six-task average); the cache's gains are in perplexity and retrieval.}
\label{tab:main}
\footnotesize
\setlength{\tabcolsep}{3pt}
\resizebox{\textwidth}{!}{%
\begin{tabular}{l cc cccccc c cc}
\toprule
 & \multicolumn{2}{c}{LM ppl $\downarrow$} & \multicolumn{7}{c}{Commonsense $\uparrow$} & \multicolumn{2}{c}{Retrieval $\uparrow$} \\
\cmidrule(lr){2-3}\cmidrule(lr){4-10}\cmidrule(lr){11-12}
Model (source) & Wiki & LMB & ARCe & ARCc & Hella & PIQA & Wino & LMBa & Avg. & FDA & SWDE \\
\midrule
Transformer++ full-attn \citep{nha2025} & 26.88 & 42.15 & 44.91 & 25.94 & 34.95 & 64.31 & 51.07 & 32.84 & 42.34 & 46.1 & 25.9 \\
\midrule
DeltaNet \citep{precond2026}            & 29.04 & 45.76 & 44.02 & 23.55 & 34.08 & 65.07 & 50.91 & 29.58 & 41.20 & 8.5 & 27.1 \\
GLA \citep{nha2025}                     & 28.78 & 39.00 & 44.53 & 22.27 & 34.84 & 63.93 & 51.38 & 32.27 & 41.54 & 11.3 & 16.8 \\
GSA \citep{nha2025}                     & 28.17 & 42.57 & 45.50 & 24.23 & 35.00 & 64.85 & 50.43 & 30.78 & 41.80 & 6.4 & 16.9 \\
KDA \citep{precond2026}                 & 26.18 & 31.37 & 45.45 & 22.70 & \textbf{36.06} & \textbf{66.00} & 52.25 & 34.04 & 42.75 & 13.9 & 34.1 \\
GDN \citep{precond2026}                 & 27.08 & 31.39 & 44.49 & 24.32 & 35.96 & 65.83 & 51.30 & 34.50 & 42.73 & 13.6 & 29.4 \\
GDN (ours, anchor)                      & 27.32 & 30.95 & \textbf{46.13} & 23.72 & 35.88 & 65.07 & 50.43 & 34.02 & 42.54 & 11.7 & 29.0 \\
\midrule
\method{}${+}$recency                   & 25.04 & 32.33 & \textbf{46.13} & \textbf{24.40} & 35.62 & 65.34 & \textbf{52.80} & \textbf{34.72} & \textbf{43.17} & 16.9 & 29.9 \\
\textbf{\method{} (ours)}               & \textbf{22.92} & \textbf{30.26} & 46.00 & 24.06 & 35.91 & 65.02 & 51.54 & 34.54 & 42.85 & \textbf{20.1} & \textbf{35.9} \\
\bottomrule
\end{tabular}}
\end{table}

\begin{itemize}[leftmargin=1.4em,itemsep=1pt,topsep=2pt]
\item \textbf{Perplexity:} \method{}'s Wiki $22.92$ is a large drop from the same-backbone GDN anchor
($27.32\!\to\!22.92$, ${-}16.1\%$), below the strongest published sub-quadratic baseline (KDA $26.18$), and even
below the full-attention Transformer++ ($26.88$); LAMBADA perplexity is also the lowest in the table ($30.26$).
\item \textbf{Commonsense:} the six-task average is effectively a tie among the strong models --- \method{} $42.85$,
the \method{}+recency control $43.17$, KDA $42.75$, and GDN ${\sim}42.7$ all sit within ${\sim}0.6$ (above Transformer++
$42.34$); commonsense is not where the cache mechanism separates the models (perplexity and retrieval are).
Over the 9-task accuracy mean, \method{} $0.446$ vs.\ GDN $0.440$ (BoolQ $0.548\!\to\!0.584$, ${+}3.5$pt;
WinoGrande, LAMBADA-acc, and HellaSwag also higher).
\item \textbf{Retrieval:} on in-context exact extraction \method{} beats \emph{all linear} baselines by a wide
margin --- FDA $11.7\!\to\!20.1$ ($+72\%$ rel.), SWDE $29.0\!\to\!35.9$ ($+24\%$), SQuAD
$32.5\!\to\!33.8$. This is the cache's intended regime: preserving exact tokens that the recurrent state would
otherwise compress away. Only the full-attention Transformer++ (FDA $46.1$) still leads pure extraction.
\end{itemize}

\subsection{Consistency across scale}
Table~\ref{tab:scale} asks whether the perplexity gain is confined to a single model size. Across 46M, 170M,
and 340M, each comparison uses a matched GDN-vs-\method{} backbone within the same scale; App.~\ref{app:scales}
lists the architecture, corpus, and context length for each run. The gain is consistent: \method{} lowers Wikitext
perplexity by $15$--$16\%$ relative to the same-backbone GDN anchor at every scale.

\begin{table}[h]
\centering
\caption{\method{} vs.\ same-backbone GDN, Wikitext perplexity across scales.}
\label{tab:scale}
\small
\begin{tabular}{lcc}
\toprule
Scale & GDN ppl $\downarrow$ & \method{} ppl $\downarrow$ \\
\midrule
46M & 71.0 & \textbf{59.5} \\
170M & 35.98 & \textbf{30.51} \\
340M & 27.32 & \textbf{22.92} \\
\bottomrule
\end{tabular}
\end{table}

\subsection{Long-Context Retrieval}
\label{sec:exp:longctx}
We evaluate on the official RULER benchmark \citep{hsieh2024ruler} (synthetic tasks, our Mistral tokenizer,
limit 100) at 2k/4k/8k/16k/32k --- up to $16\times$ the 2k training length. On the headline single
needle-in-a-haystack task, S-NIAH-1, \textbf{the recurrent state (GDN) collapses with length while \method{}
stays robust}, with a matched \method{}+recency cache in between (Figure~\ref{fig:teaser}b): at 32k, \method{}
reaches $0.58$ vs.\ \method{}+recency $0.24$ vs.\ GDN $0.14$ --- a $+0.44$ margin over the state, and $+0.64$ at 16k.

The advantage extends to most above-floor RULER cells (Table~\ref{tab:sniah}), grouped into single-needle
(S-NIAH-1/2/3) and multi-needle (multi-key/-value/-query): \method{} wins the harder single-needle variants by
wide margins where they are not floored (S-NIAH-2 at 8k, $0.35$ vs.\ $0.09$) and most multi-needle cells
(multi-value $0.28$ vs.\ $0.17$ at 2k), with a few near-tie exceptions. We also include a full-attention
Transformer++ as a 2k full-attention ceiling: within its training length it is the strongest model (exact
softmax retrieval), but this RoPE checkpoint is not a length-extrapolating baseline and every shown task drops
to $0$ at 4k and beyond --- exactly the regime where \method{} remains useful.

\begin{table}[h]
\centering
\caption{RULER accuracy (340M), compact 2k--8k multi-task snapshot. Transformer++ is the
full-attention ceiling (RoPE, max position 8192); we compare three sub-quadratic models (GDN / \method{}${+}$recency /
\method{}). Single-needle $=$ S-NIAH-1/2/3 (columns 1--3); multi-needle $=$ multi-key-1 (MK1), multi-value
(MV), multi-query (MQ). Bold $=$ best \emph{sub-quadratic} model per cell. Transformer++ is strongest within
its 2k training length but collapses to $0$ under
RoPE extrapolation at 4k$+$, where \method{} degrades gracefully. The 16k/32k S-NIAH-1 trend is shown in
Fig.~\ref{fig:teaser}b.}
\label{tab:sniah}
\small
\setlength{\tabcolsep}{5pt}
\begin{tabular}{ll cccccc}
\toprule
 & & \multicolumn{3}{c}{Single-needle} & \multicolumn{3}{c}{Multi-needle} \\
\cmidrule(lr){3-5}\cmidrule(lr){6-8}
len & model & 1 & 2 & 3 & MK1 & MV & MQ \\
\midrule
\multirow{4}{*}{2k} & Transformer++ & 1.00 & 1.00 & 0.79 & 0.71 & 0.34 & 0.33 \\
                    & GDN & 1.00 & 0.38 & 0.26 & 0.17 & 0.17 & \textbf{0.24} \\
                    & \method{}${+}$recency & 1.00 & 0.83 & 0.89 & \textbf{0.26} & 0.21 & 0.22 \\
                    & \method{} & 1.00 & \textbf{1.00} & \textbf{0.96} & 0.25 & \textbf{0.28} & 0.18 \\
\midrule
\multirow{4}{*}{4k} & Transformer++ & 0.00 & 0.00 & 0.00 & 0.00 & 0.00 & 0.00 \\
                    & GDN & \textbf{1.00} & 0.52 & 0.23 & 0.17 & 0.14 & 0.19 \\
                    & \method{}${+}$recency & 0.93 & 0.27 & 0.38 & 0.13 & 0.17 & 0.17 \\
                    & \method{} & 0.99 & \textbf{0.89} & \textbf{0.43} & \textbf{0.30} & \textbf{0.28} & \textbf{0.26} \\
\midrule
\multirow{4}{*}{8k} & Transformer++ & 0.00 & 0.00 & 0.00 & 0.00 & 0.00 & 0.00 \\
                    & GDN & 0.83 & 0.09 & \textbf{0.07} & 0.03 & 0.07 & 0.02 \\
                    & \method{}${+}$recency & 0.74 & 0.04 & 0.02 & 0.05 & 0.03 & 0.01 \\
                    & \method{} & \textbf{0.98} & \textbf{0.35} & 0.05 & \textbf{0.11} & \textbf{0.16} & \textbf{0.08} \\
\bottomrule
\end{tabular}
\end{table}

\subsection{Ablations}
\textbf{\method{}+recency vs.\ importance eviction (340M, matched memory).} The sharpest test of contribution~1 is a
matched control --- the identical architecture ($w{=}64$, same chunk, sharpened read, gate, and kernel),
changing \emph{only} the eviction signal: \emph{position} (keep the most recent $w$) vs.\ \emph{surprise}
(keep top-$w$ by $\be$). We report the \method{}+recency control as a first-class row throughout, and importance wins on the axes
the cache is designed to affect: lower perplexity (Wiki $25.04\!\to\!22.92$; Table~\ref{tab:main}) and far stronger
long-context retrieval (Figure~\ref{fig:teaser}b, Table~\ref{tab:sniah}; S-NIAH-1 at 32k, \method{}+recency $0.24$ vs.\
\method{} $0.58$). Most tellingly, \textbf{the recency cache barely improves on no cache at all on the far
needle}: at 32k it recalls only $0.24$, marginally above the no-cache state (GDN $0.14$) and far below importance
eviction (\method{} $0.58$) --- a recency window gives little far-needle benefit, because the needle slides out
of it, whereas surprise-based eviction keeps it. For a bounded exact memory, \emph{what} to cache matters more
than \emph{how recent}.\footnote{For the recency control, we use full recomputation to avoid implementation-path
confounds; this is the path used for the PPL, commonsense, retrieval, and RULER numbers we report.}

\textbf{Eviction signal} (Table~\ref{tab:evict}). We isolate the eviction rule with the same 46M backbone,
cache size, and flat cache read, and evaluate the two axes a bounded cache is meant to balance: whether the
state still carries a far needle after the cache window has moved on, and how many exact key--value facts the
cache can retain within its span. The far-needle column uses a lightweight passkey-style probe inspired by
\citet{mohtashami2023landmark}: we place a numeric key at a controlled depth in a 4k context, teacher-force
the answer tokens after the query, and report next-token accuracy. Pure residual surprise, $\norm{e}$, is not
enough; it lacks the write-strength utility signal and performs poorly on the far needle. Multiplying by the
GDN write gate gives $\be$, the actual delta-rule update magnitude, which is best or tied-best on every
diagnostic column while also giving the lowest WikiText perplexity.

\begin{table}[h]
\centering
\caption{Eviction-signal diagnostic (46M, flat read). Far is teacher-forced passkey accuracy at depth
$0.1$ in a 4k context; multi-key columns report associative-recall capacity for 1/2/4 facts; perplexity is
WikiText-103 test.}
\label{tab:evict}
\footnotesize
\setlength{\tabcolsep}{2.8pt}
\begin{tabular}{llccccc}
\toprule
Eviction score & has $\beta$? & far $d{=}0.1$ & 1 key & 2 keys & 4 keys & Wiki PPL $\downarrow$ \\
\midrule
GDN (no cache) & -- & 0.55 & 0.78 & 0.45 & 0.41 & 70.21 \\
cumulative attention (H2O) & -- & 0.43 & \textbf{0.97} & 0.69 & 0.55 & 71.45 \\
residual $\norm{e}$ & no & 0.22 & 0.71 & 0.61 & 0.48 & 70.50 \\
$\beta\norm{v}$ & yes & 0.42 & \textbf{0.97} & 0.72 & 0.54 & 70.76 \\
\textbf{$\be$ (ours)} & yes & \textbf{0.67} & \textbf{0.97} & \textbf{0.74} & \textbf{0.56} & \textbf{70.10} \\
\bottomrule
\end{tabular}
\end{table}

\textbf{Sharpened read} (Table~\ref{tab:sharp}). With the eviction rule fixed to $\be$, the decisive read-side
change is not a larger fixed temperature but the normalization used to form cache logits. Unit-L2 queries and
keys make the cache read too flat, so the cache behaves like a soft average. RMSNorm-$\gamma$ keeps the natural
$\sqrt{d}$ logit scale while letting the model tune it, turning the same bounded cache into an exact local memory:
it lowers WikiText perplexity by more than ten points and sharply improves multi-key capacity, without losing the
far-needle behavior of the recurrent state.

\begin{table}[h]
\centering
\caption{Cache-read normalization (46M). The GDN and unit-L2 rows repeat the flat-read anchors from
Table~\ref{tab:evict} where columns overlap; the RMSNorm-$\gamma$ row is a three-seed mean. Far is
teacher-forced passkey accuracy at depth $0.1$ in a 4k context; mean is the passkey average over depths;
capacity reports multi-key associative recall.}
\label{tab:sharp}
\footnotesize
\setlength{\tabcolsep}{2.2pt}
\begin{tabular}{lccccccc}
\toprule
Read / model & Wiki PPL $\downarrow$ & far $d{=}0.1$ & passkey mean & 1 key & 4 keys & 8 keys & 16 keys \\
\midrule
GDN (no cache) & 70.21 & 0.55 & 0.60 & 0.78 & 0.41 & 0.35 & 0.31 \\
$\be$ + unit-L2 read & 70.10 & 0.67 & 0.67 & 0.97 & 0.56 & 0.41 & 0.31 \\
\textbf{$\be$ + RMSNorm-$\gamma$ (ours)} & \textbf{59.5} & \textbf{0.75} & \textbf{0.69} & \textbf{0.97} & \textbf{0.77} & \textbf{0.67} & \textbf{0.41} \\
\bottomrule
\end{tabular}
\end{table}

%% file: sec_conclusion.tex
\section{Conclusion and Limitations}
\label{sec:conclusion}

\paragraph{Conclusion.}
Starting from a semiparametric test-time regression view --- a linear-attention recurrent state is the
parametric estimator for compressible key--value structure, but needs a bounded set of exact KV pairs for
exact associations --- we attached a hippocampus-like KV cache to every layer of Gated DeltaNet. The cache is
guided by two design choices: (i) eviction driven by a \emph{parameter-free, intrinsic} surprise signal $\be$
(the delta-rule write magnitude) --- keep what wrote most to the state, not what is most recent; and (ii) a read
sharpened by a \emph{decoupled} RMSNorm-$\gamma$, so the cache performs exact retrieval rather than soft
averaging. At 340M parameters trained on 15B SlimPajama tokens, \method{} improves language modeling and recall together: lowest perplexity
of all compared models (Wikitext 22.92, below a full-attention Transformer++), best linear in-context retrieval,
and competitive commonsense (six-task average); it is length-robust on RULER needle-in-a-haystack recall out to 32k
($16\times$ the training length), holding $0.58$ where the recurrent baseline collapses to $0.14$ and the
2k-trained full-attention checkpoint reaches $0$.
The perplexity advantage holds from 46M to 340M. The take-away: \emph{a linear-attention model's own update rule already
diagnoses what it fails to remember; spending a small exact memory on exactly those ``surprising'' tokens
recovers the long-range exact recall it loses.}

\paragraph{Limitations.}
The cache is deliberately bounded: it spans only $w{+}C{+}1{\approx}321$ tokens, so in very long or needle-dense contexts it cannot retain every relevant item, and single-needle recall is $0.58$ rather than perfect at 32k. \method{} also narrows, but does not close, the gap to full attention on pure token-exact extraction such as FDA, where every token can matter. Finally, the main-scale results are single-seed up to 340M; while the matched recency comparison and 46M diagnostics support the $\be$ eviction rule, we have not run a matched-memory comparison against learned eviction modules such as LTE's CNN.

%% file: sec_appendix.tex
\section{Scale configurations}
\label{app:scales}
Table~\ref{tab:scaleconfig} lists the architecture, corpus size, and context length for the scaling comparison in
Table~\ref{tab:scale}. Within each row, GDN and \method{} use the same backbone and training recipe; \method{}
only adds the bounded exact KV cache.

\begin{table}[h]
\centering
\caption{Scale configurations for Table~\ref{tab:scale}.}
\label{tab:scaleconfig}
\small
\begin{tabular}{lccccc}
\toprule
Scale & $d_{\mathrm{model}}$ & layers & corpus & train tokens & ctx \\
\midrule
46M & 512 & 12 & FineWeb-Edu & 0.5B & 4096 \\
170M & 1024 & 12 & SlimPajama & 6.22B & 2048 \\
340M & 1024 & 24 & SlimPajama & 15.0B & 2048 \\
\bottomrule
\end{tabular}
\end{table}

For 170M and 340M, the architecture family follows the GDN recipe: $4$ heads $\times$ head-dim $256$,
$\mathrm{expand\_v}{=}1$, $\mathrm{hidden\_ratio}{=}4$, conv 4, tied embeddings, vocabulary $32000$, Mistral
tokenizer, and AdamW with peak lr $4{\times}10^{-4}$. The 46M row uses a smaller 12-layer, $d_{\mathrm{model}}{=}512$ architecture trained on
FineWeb-Edu for 0.5B tokens at ctx 4096; this is the scale used for the component studies in
Tables~\ref{tab:evict}--\ref{tab:sharp}.

\textbf{\method{} cache hyperparameters:} \texttt{evict=betae, window=64, chunk=256, gate\_init=$-4.0$,
cache\_norm=rms, tau\_init=1.0, tau\_freeze=true, cache\_kernel=sdpa}. The GDN baseline shares all other
settings within each row.